\pdfoutput=1

\documentclass[11pt]{article}

\usepackage[]{EMNLP2022}

\usepackage{times}
\usepackage{latexsym}

\usepackage[T1]{fontenc}

\usepackage[utf8]{inputenc}

\usepackage{microtype}

\usepackage{inconsolata}

\usepackage{xcolor}
\usepackage{pgfplots}
\usepackage{tikz}

\usepackage{siunitx}

\PassOptionsToPackage{linktocpage}{hyperref}

\definecolor{bblue}{HTML}{4F81BD}
\definecolor{rred}{HTML}{C0504D}

\title{University of Cape Town's WMT22 System: Multilingual Machine Translation for Southern African Languages}

\author{Khalid N.~Elmadani \, Francois Meyer \, Jan Buys\\
  Department of Computer Science \\
  University of Cape Town \\
  \texttt{\{ahmkha009,myrfra008\}@myuct.ac.za, jbuys@cs.uct.ac.za}}

\begin{document}
\maketitle

\begin{abstract}
The paper describes the University of Cape Town's submission to the constrained track of the WMT22 Shared Task: Large-Scale Machine Translation Evaluation for African Languages. Our system is a single multilingual translation model that translates between English and 8 South / South East African Languages, as well as between specific pairs of the African languages. 
We used several techniques suited for low-resource machine translation (MT), including overlap BPE, back-translation, synthetic training data generation, and adding more translation directions during training. Our results show the value of these techniques, especially for directions where very little or no bilingual training data is available.\footnote{Our model is available at \href{https://github.com/Khalid-Nabigh/UCT-s-WMT22-shared-task}{https://github.com/Khalid-Nabigh/UCT-s-WMT22-shared-task}.}

\end{abstract}

\section{Introduction}

Southern African languages are underrepresented in NLP research, in part because most of them are low-resource languages: It is not always possible to find high-quality datasets that are large enough to train effective deep learning models \citep{kreutzer-etal-2021-quality}. The WMT22 Shared Task on Large-Scale Machine Translation Evaluation for African Languages \citep{adelani-etal-2022-findings} presented an opportunity to apply one of the most promising recent developments in NLP --- multilingual neural machine translation --- to Southern African languages. For many languages, the parallel corpora released for the shared task are the largest publicly available datasets yet. For some translation directions (e.g. between Southern African languages), no parallel corpora were previously available. 

In this paper we present our submission to the shared task. Our system is a Transformer-based encoder-decoder \citep{vaswani-etal-2017-attention} that translates between English and 8 South / South East African languages (Afrikaans, Northern Sotho, Shona, Swati, Tswana, Xhosa, Xitsonga, Zulu) and in 8 additional directions (Xhosa to Zulu, Zulu to Shona, Shona to Afrikaans, Afrikaans to Swati, Swati to Tswana, Tswana to Xitsonga, Xitsonga to Northern Sotho, Northern Sotho to Xhosa).
We trained a single model with shared encoder and decoder parameters and a shared subword vocabulary.

We applied several methods aimed at improving translation performance in a low-resource setting.
We experimented with BPE \citep{sennrich-etal-2016-neural} and overlap BPE \citep{patil-etal-2022-overlap}, the latter of which increases the representation of low-resource language tokens in the shared subword vocabulary. 
We used initial multilingual and bilingual models to generate back-translated sentences \citep{sennrich-etal-2016-improving} for subsequent training.


First, we trained a model to translate between English and the 8 Southern African languages.  Then we added the 8 additional translation directions and continued training. For some of these additional directions no parallel corpora were available, so we generated synthetic training data with our existing model. By downsampling some of the parallel corpora to ensure a balanced dataset, we were able to train our model effectively in the new directions, while retaining performance in the old directions.


We describe the development of our model and report translation performance at each training stage. Our final results compare favourably to existing works with overlapping translation directions. While there is considerable disparity in performance across languages, our model nonetheless achieves results that indicate some degree of effective MT across all directions (most BLEU scores are above 10 and most chrF++ scores are above 40). 
We also discuss our findings regarding techniques for low-resource MT. We found overlap BPE and back-translation to improve performance for most translation directions. 
Furthermore, our results confirm the value of multilingual models, which proves critical for the lowest-resource languages.

\section{Background}

\subsection{Multilingual Neural Machine Translation (MNMT)}
Multilingual models help low-resource languages (LRLs) by leveraging the massive amount of training data available in high-resource languages (HRLs) \citep{aharoni-etal-2019-massively,zhang-etal-2020-improving}. In the context of Neural Machine Translation, a multilingual model can translate between more than two languages. Current research in MNMT can be divided into two main areas: training language-specific parameters \citep{kim-etal-2019-effective,philip-etal-2020-monolingual} and training a single massive model that shares all parameters among all languages \citep{beyond-english,nllb}. Our work lies in the second category, as we are building a single multilingual translation system by exploring back-translation and different vocabulary generation approaches.

\subsection{Back-Translation}
Given parallel sentences in two languages $A$ and $B$ ($A_b$, $B_a$), with goal of training a model that translates sentences from $A$ to $B$ ($A \to B$). Back-translation works as follows: First, one trains a ($B\,\to\,A$) model using the available ($A_b$, $B_a$) data. Then the $B_a$ sentences are passed to the model to regenerate $A_b$. This model's output ($A_b^{\prime}$) is then considered as additional synthetic parallel data ($A_b^{\prime}$, $B_a$). The final step of back-translation is training an ($A \to B$) translation model using ($A_b^{\prime}$, $B_a$) as parallel data. The motivation behind back-translation is that the noise added to the $A_b^{\prime}$ sentences from regeneration increases the model's robustness \citep{edunov-etal-2018-understanding-back}. The same approach can be extended to multilingual models \citep{liao-etal-2021-back-wmt}.

\subsection{Overlap-based BPE (OBPE)}
Byte Pair Encoding (BPE) is a vocabulary creation method that relies on $n$-gram frequency \citep{sennrich-etal-2016-neural}. The starting point is a character-based vocabulary. At each step, the BPE algorithm identifies the two adjacent tokens with the highest frequency, joins them together as a single token, and adds the new token to the vocabulary. The dataset is then restructured based on the expanded vocabulary. In the case of multilingual training, a single BPE vocabulary can handle all languages by running the BPE algorithm on the union of the data from all the languages. However, when constructing a multilingual vocabulary, BPE will prefer frequent word types, most of which are from HRLs, leaving a smaller proportion of the vocabulary for words from LRLs. 

Overlap-based BPE (OBPE) is a modification to the BPE vocabulary creation algorithm which enhances overlap across related languages \citep{patil-etal-2022-overlap}. OBPE takes into account the frequency of tokens as well as their existence among different languages. Given a list of HRLs ($L_{\mathrm{HRL}}$) and LRLs ($L_{\mathrm{LRL}}$), OBPE tries to balance cross-lingual sharing (tokens shared between HRLs and LRLs) and individual languages’ representation. The optimal OBPE vocabulary for a set of languages from different families is produced by considering the highest resource language from each family as $L_{\mathrm{HRL}}$ and the rest of the languages as $L_{\mathrm{LRL}}$.

\begin{table}
\centering
\begin{tabular}{ll}
\hline
\textbf{Language Pairs} & {\bf \texttt{WMT22\_african}}\\
\hline
eng-sna & 8.7M \\
eng-xho & 8.6M \\
eng-tsn & 5.9M \\
eng-zul & 3.8M \\
eng-nso & 3M \\
eng-afr & 1.6M \\
eng-tso & 630K \\
eng-ssw & 165K \\
\hline
xho-zul & 1M \\
zul-sna & 1.1M \\
sna-afr & 1.6M* \\
afr-ssw & 165K* \\
ssw-tsn & 85K \\
tsn-tso & 285K \\
tso-nso & 212K \\
nso-xho & 200K \\
\hline
\end{tabular}
\caption{Number of available parallel sentences for all language pairs. * indicates that no data is available for these pairs and the number represents the amount of synthetic data we generated.}
\label{tab:dataset}
\end{table}

\begin{table}
\centering
\resizebox{0.48\textwidth}{!}{\begin{tabular}{lll}
\hline
\textbf{Language Family} & $\mathbf{L_{HRL}}$ & $\mathbf{L_{\mathrm{LRL}}}$\\
\hline
Germanic & English(eng) & Afrikaans(afr) \\
Nguni & Xhosa(xho) & Zulu(zul), Swati(ssw) \\
Sotho-Tswana & Tswana(tsn) & Sepedi(nso) \\
Bantu & Shona(sna) & Xitsonga(tso) \\
\hline
\end{tabular}}
\caption{The languages included in our translation system, grouped by language family and whether they are used as $L_{\mathrm{HRL}}$ or $L_{\mathrm{LRL}}$ for the OBPE algorithm.}
\label{tab:families}
\end{table}

\section{Datasets}
The WMT22 dataset is released along with the shared task. It contains bitext for 248 pairs of African languages, referred to as  \texttt{WMT22\_african}.\footnote{\url{https://huggingface.co/datasets/allenai/wmt22_african}} We use \texttt{WMT22\_african} for both training and validation; the first 3 000 sentences from each language pair is reserved for validation and the rest for training. Table \ref{tab:dataset} shows available number of sentences for each language pair. No data was provided for Shona-Afrikaans and Afrikaans-Swati, so we generated synthetic data for these translation directions (see section \ref{sec:syn}). For testing, we used the Flores dev set, which contains 997 parallel sentences for each language pair. Additionally, we report the results of the final translation system as evaluated by the shared task organizers on a hidden test set. 

\subsection{OBPE}
We trained BPE and OBPE tokenizers using the $eng \leftrightarrow \mathrm{LRL}$ data only (the first 8 rows of table \ref{tab:dataset}). The vocabulary size for both BPE and OBPE is set to 40K. For OBPE, the $L_{\mathrm{HRL}}$ contains the highest-resource language from each language family ($eng, xho, tsn, sna$), while $L_{\mathrm{LRL}}$ includes the rest of the languages (see table \ref{tab:families}). We used \citeposs{patil-etal-2022-overlap} implementation for both BPE and OBPE. This implementation is based on the Hugging Face Tokenizers library.\footnote{\url{https://huggingface.co/docs/tokenizers}}

\begin{figure*}
\centering
\begin{tikzpicture}
    \begin{axis}[
        width  = \textwidth,
        height = 8cm,
        major x tick style = transparent,
        ybar,
        bar width=12pt,
        ymajorgrids = true,
        ylabel = {\% change in number of training tokens from BPE to OBPE},
        symbolic x coords={eng,sna,xho,tsn,zul,nso,afr,tso,ssw},
        xtick = data,
        scaled y ticks = false,
    ]
        \addplot[style={bblue,fill=bblue,mark=none}]
            coordinates {(eng, 1.17) (sna, -1.03) (xho, -0.90) (tsn, 0.24) (zul, -2.29) (nso, -0.88) (afr, -2.18) (tso, -0.94) (ssw, -0.96)};

    \end{axis}
\end{tikzpicture}
\caption{The change in the number of tokens in the training set per language when using OBPE instead of BPE. Less training tokens correspond to better a representation of a language in the shared subword vocabulary, so negative percentage changes reflect an improvement in low-resource language representation.} 
\label{fig:representation}
\end{figure*}
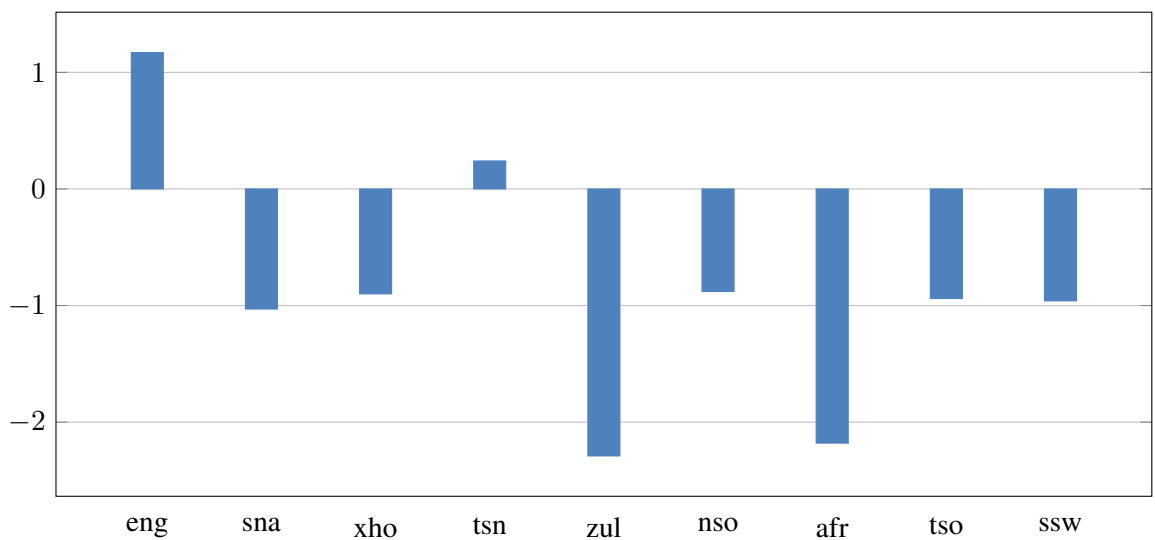

\begin{figure*}
\centering
\begin{tikzpicture}
    \begin{axis}[
        width  = \textwidth,
        height = 8cm,
        major x tick style = transparent,
        ybar,
        bar width=12pt,
        ymajorgrids = true,
        ylabel = {Average no. tokens per sentence pairs},
        symbolic x coords={eng-sna,eng-xho,eng-tsn,eng-zul,eng-nso,eng-afr,eng-tso,eng-ssw},
        xtick = data,
        scaled y ticks = false,
    ]
        \addplot[style={bblue,fill=bblue,mark=none}]
            coordinates {(eng-afr, 37.34) (eng-nso, 25.78) (eng-sna, 34.19) (eng-ssw, 29.11) (eng-tsn, 26.68) (eng-xho, 34.33) (eng-tso, 34.39) (eng-zul, 35.75)};

        \addplot[style={rred,fill=rred,mark=none}]
             coordinates {(eng-afr, 37.11) (eng-nso, 25.77) (eng-sna, 34.20) (eng-ssw, 29.09) (eng-tsn, 26.84) (eng-xho, 34.38) (eng-tso, 34.35) (eng-zul, 35.51)};

        \legend{BPE,OBPE}
    \end{axis}
\end{tikzpicture}
\caption{The average number of tokens per sentence pair for all language pairs with English, comparing BPE and OBPE vocabularies. More tokens lead to slower training.}
\label{fig:speed}
\end{figure*}
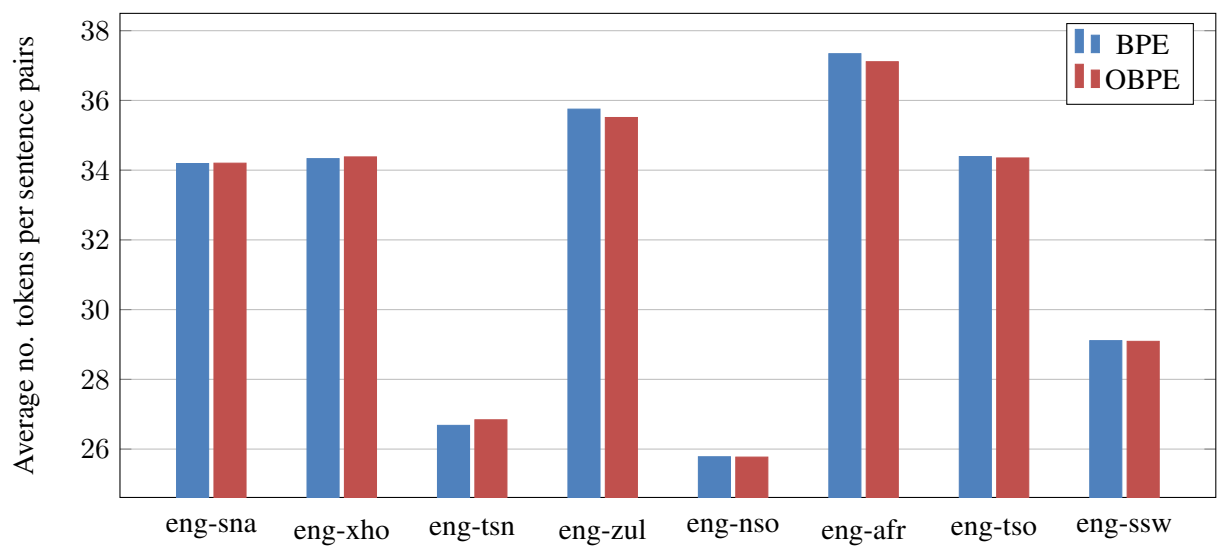

\section{Methodology}
In this work, we only focus on South and South East African languages, their translation to/from English, and eight translation directions between these languages. We divided the training of the translation system into two stages. In the first stage, we trained a multilingual model for translating from all LRLs to English and vice versa. To incorporate the translation directions between LRLs into the system, we did further training on the translation model from stage 1. We divided the training process into stages instead of training the model in one session due to computational resource constraints. Both stages are explained in more detail below.

All models were trained with the \texttt{Fairseq} toolkit \citep{ott2019fairseq}. We used the \texttt{transformer-base} architecture \citep{vaswani-etal-2017-attention} for training all bilingual models. We base the multilingual models on the \texttt{BART} architecture \citep{liu-etal-2020-mbart-multilingual-denoising}, using \citeposs{tang-etal-2021-fb-multilingual} implementation and hyperparameters, including adding a token to indicate the source language before the input sentence and a token for the target language before the output sentence.

\begin{table}
\centering
\begin{tabular}{c|S[table-format=3.2]}
\hline
\textbf{Data} & $\mathbf{\Delta}$ \\
\hline
sna-eng & 0.1 \\
xho-eng & 0.2 \\
tsn-eng & -0.2 \\
zul-eng & -0.7 \\
nso-eng & 0.3 \\
afr-eng & 0.0 \\
tso-eng & 0.0 \\
ssw-eng & 0.3 \\
\hline
eng-sna & 0.1 \\
eng-xho & -0.2 \\
eng-tsn & 0.2 \\
eng-zul & 0.1 \\
eng-nso & -0.2 \\
eng-afr & 0.0 \\
eng-tso & -0.2 \\
eng-ssw & 0.0 \\
\hline
\end{tabular}
\caption{BLEU score differences between the OBPE multilingual model (13th epoch) and the BPE multilingual model (10th epoch) on Flores dev set.
We stopped training the BPE model at this point as the OBPE model is computationally more efficient. The translation directions are sorted based on the available amount data.}
\label{tab:res-method}
\end{table}

\subsection{Stage 1: Translation Between LRLs and English}
We used BPE and OBPE vocabularies to train two multilingual models for all directions between English and LRLs. 
Bilingual models were trained for each translation direction using a single vocabulary for each model. Finally, we performed back-translation for all directions using the model with the highest BLEU score in each case.

\subsubsection{Multilingual Training}
Multilingual models generally have more parameters and require more training time and computational resources than bilingual models. 
Computational constraints prevented us from fully training two multilingual models and then doing back-translation from them. Subsequently we used BPE and OBPE vocabularies to train two multilingual models till the 10th and 13th epochs, respectively. At this point, we found that the difference in translation quality between the two models is negligible (see table \ref{tab:res-method}). However, the OBPE model is slightly faster in training and represent LRLs better.
A language $l$ is represented better in vocabulary $V_1$ than $V_2$ if $V_1$ contains more subword tokens from $l$ than $V_2$. The total number of tokens in $l$’s training data will influence its representation in the vocabulary. Reducing the number of tokens in the training sentences requires increasing the vocabulary capacity. Therefore, fewer tokens in the training data corresponds to a better vocabulary representation. We are interested in comparing BPE and OBPE's vocabulary representation for all languages. We used the following formula to measure the relative change in the number of training tokens when using OBPE instead of BPE, 
\begin{equation}
\textrm{change}_l=\frac{T_{\textrm{OBPE}}^l-T_{\textrm{BPE}}^l}{T_{\textrm{BPE}}^l} \%
\end{equation}
where $T_{\textrm{BPE}}^l$ and $T_{\textrm{OBPE}}^l$ are the total number of tokens in language $l$’s training data when using BPE and OBPE vocabularies, respectively. Figure \ref{fig:representation} shows the change in number of training tokens for all languages. The negative sign in the figure indicates that OBPE represents the language better than BPE. It can be clearly seen that OBPE represents most LRLs better than BPE.

As we are training autoregressive models, the training speed depends on the number of target tokens, which is controlled by the target language representation in the subword vocabulary. 
Therefore we use the average number of tokens per training example for each language pair (\emph{eng}-\emph{l}) as a proxy for training speed.
Fewer tokens leads to faster training. Both source and target tokens are included, as we are training the model to translate in both directions:

\begin{equation}
\textrm{AvgTokens}_{eng-l}^V=\frac{\textrm{Tok}_{eng-l}^l+\textrm{Tok}_{eng-l}^{eng}}{N_{eng-l}}
\end{equation}
where $\textrm{AvgTokens}_{eng-l}^V$ indicates the average number of tokens in one training example from the $eng-l$ dataset using $V$ vocabulary. $\textrm{Tok}_{eng-l}^l$ and $\textrm{Tok}_{eng-l}^{eng}$ represent the total of $l$ and $eng$ tokens, respectively, in the $eng-l$ dataset, while $N_{eng-l}$ represents the number of training examples in the same dataset. Figure \ref{fig:speed} shows the average number of training tokens in each language pair when using BPE and OBPE vocabularies. We observe that training with OBPE is slightly faster than training with BPE. The speed difference is higher for languages that are better represented by OBPE (see figure \ref{fig:representation}).

For these two reasons, and due to time and resources constrains, we chose to continue with training the OBPE multilingual model only.

\subsubsection{Bilingual Training}
Multilingual models often harm performance on high-resource languages compared to their bilingual counterparts \citep{ijcai2022p0619}.
For back-translation, we used bilingual models for the subset of language pairs where this happens. We had two translation directions for each language (from/to English) and two vocabulary options (BPE/OBPE) for each direction. We ended up with 32 bilingual models.

All bilingual models were trained on either an \texttt{Nvidia} A100 full card (40GB) or a division of half a card (20GB) for 45 epochs with a batch size of 12 288 tokens. The training time depends on the language pairs, but the highest-resource language pair took three days of training.

\subsubsection{Back-Translation}
For each translation direction, we choose one of the following models for generating back-translation sentences: OBPE bilingual, BPE bilingual, and the 17th epoch checkpoint from the OBPE multilingual model. The selection is based on the models’ performance on the Flores dev set, as measured by their BLEU score. We generated the back-translation sentences from the available parallel data only; no additional monolingual data was used. Results from table \ref{tab:results} show the performance of those three models. It can be seen that bilingual models are performing better in both directions of the higher-resource language pairs and for \emph{eng-afr}. We discuss the results in more details in section \ref{sec:res}.

We trained the OBPE multilingual model until the 17th epoch. That checkpoint was then used to generate back-translation data for the directions where the multilingual models outperform bilingual ones. Due to resources and time constraints, we started training the back-translation multilingual model from the 17th epoch checkpoint of the OBPE multilingual model. The OBPE multilingual model continued training regularly from the 17th epoch.

We ran all multilingual experiments on 2 \texttt{Nvidia} A100 cards (40GB each). One epoch of back-translation or OBPE multilingual models took 16 hours. Both models trained for 45 epochs with a batch size of 16 384 tokens, leading to a total training time of 30 days for each model.

After training both multilingual models, we had four models for each translation direction; two bilingual and two multilingual models.

\subsection{Stage 2: Translation Between LRLs}

At this stage we found that our models showed adequate performance in the English-centric directions (similar evaluation scores to existing works with overlapping translation directions). The goal of the next stage was to add new translation directions between specific LRLs. 
Our best multilingual model at this point (based on BLEU scores in the English-centric directions) was the OBPE-based model that was partially trained on back-translated data. Therefore we selected this model to continue training in the new directions. The model trained for an additional 39 epochs on a training set covering the old and new directions (details in section \ref{sec:bal}). This took 9 days on a full \texttt{Nvidia} A100  card (40GB), at which point validation performance had stopped improving. This resulting model is the system we submitted to the shared task.

\subsubsection{Synthetic training data} \label{sec:syn}

As shown in table \ref{tab:dataset}, the translation directions between LRLs (new directions) generally had smaller datasets than the directions from/to English (old directions). In fact, two of the new directions (Shona to Afrikaans and Afrikaans to Swati) had no parallel corpora at all. To add these two directions to the model, we generated partially synthetic training data using the available English-centric parallel corpora. Using our multilingual model, we translated the English sentences in the English-Afrikaans corpus to Shona, and the English sentences in the English-Siswati corpus to Afrikaans. This produced parallel corpora for Shona-Afrikaans and Afrikaans-Siswati, where the target sentences were real and the source sentences were synthetic. 

\subsubsection{Balancing parallel corpora} \label{sec:bal}

The challenge in adding new translation directions is to strike a balance between gaining performance in the new directions, while ensuring that performance in the old directions does not deteriorate in the process. For this stage our model was trained on parallel corpora in the old and new directions.
Including training data for the old directions ensures that the model does not lose its translation abilities for these directions. However, the parallel corpora for the old directions are on average much larger than those of the new directions. Therefore training on such an unbalanced dataset would likely result in suboptimal performance for new directions.

To counter this, we downsampled the training data for the old directions to match the corresponding corpora in the new directions in order to balance the model's exposure to the old and new directions during training. 
For example, to balance Xhosa to Zulu training (1M sentences), we trained on 1M sentences only from both the English to Zulu and the Xhosa to English corpora. 
Therefore the encoder is trained for Xhosa balancing the Xhosa-English and Xhosa-Zulu data, while the decoder is trained for Zulu balancing the English-Zulu and Xhosa-Zulu setting.

Another potentially better approach is upsampling the training data for new directions. This technique would ensure that the model is exposed to all training data of old directions. However, we did not explore this due to time constraints.




\begin{table*}
\centering
\resizebox{\textwidth}{!}{\begin{tabular}{c|S[table-format=3.2]S[table-format=3.2]|S[table-format=3.2]|S[table-format=3.2]S[table-format=3.2]|S[table-format=3.2]}
\hline
\textbf{Data} & \textbf{Bi-BPE} & \textbf{Bi-OBPE} & \textbf{M-OBPE@17} & \textbf{M-OBPE} & \textbf{M-OBPE+back} & \textbf{M-OBPE-final}\\
\hline
sna-eng & 19.1 & $\underline{19.6}$ & 17.7 & 19.1 & 18.1 & $\mathbf{19.5}$ \\
xho-eng & 26.2 & $\underline{26.9}$ & 24.2 & 26.3 & 26.5 & $\mathbf{27.5}$ \\
tsn-eng & 11.8 & 11.9 & $\underline{18.1}$ & 19.2 & 16.1 & $\mathbf{20.3}$ \\
zul-eng & $\underline{28.7}$ & 28.2 & 26.4 & 28.6 & $\mathbf{30.0}$ & $\mathbf{30.0}$ \\
nso-eng & 12.9 & 14.6 & $\underline{23.1}$ & 25.5 & 22.9 & $\mathbf{26.9}$ \\
afr-eng & 47.5 & $\mathbf{\underline{48.5}}$ & 41.8 & 45.0 & 46.4 & 44.8 \\
tso-eng & 1.1 & 3.3 & $\underline{17.2}$ & 18.8 & 16.9 & $\mathbf{20.7}$ \\
ssw-eng & 0.7 & 0.9 & $\underline{19.4}$ & 21.3 & 18.0 & $\mathbf{23.0}$ \\
\hline
avg & 18.5 & 19.2 & 23.5 & 25.5 & 24.4 & $\mathbf{26.6}$ \\
\hline
eng-sna & $\underline{10.1}$ & 9.9 & 9.3 & 10.0 & 10.1 & $\mathbf{10.3}$ \\
eng-xho & 12.3 & $\underline{12.6}$ & 10.9 & 11.8 & $\mathbf{12.7}$ & 12.1 \\
eng-tsn & 10.2 & 9.6 & $\underline{16.5}$ & 17.8 & 17.8 & $\mathbf{18.2}$ \\
eng-zul & $\underline{14.9}$ & 14.3 & 12.6 & 14.2 & $\mathbf{15.1}$ & 15.0 \\
eng-nso & 9.8 & 10.4 & $\underline{20.3}$ & 22.1 & 22.3 & $\mathbf{23.1}$ \\
eng-afr & $\mathbf{\underline{37.2}}$ & 35.8 & 32.3 & 34.1 & 36.2 & 35.6 \\
eng-tso & 0.7 & 0.9 & $\underline{12.8}$ & 14.5 & 15.0 & $\mathbf{16.9}$ \\
eng-ssw & 0.7 & 0.9 & $\underline{6.2}$ & 6.9 & 7.0 & $\mathbf{7.7}$ \\
\hline
avg & 12 & 11.8 & 15.1 & 16.4 & 17 & $\mathbf{17.4}$ \\
\hline
\end{tabular}}
\caption{BLEU scores on Flores dev set for translating between English and LRLs. The translation directions are sorted based on the available amount data. Bi-BPE and Bi-OBPE are the BPE and OBPE bilingual models, respectively. M-OBPE@17 is the 17th epoch checkpoints of the OBPE multilingual model, while M-OBPE is trained for 45 epochs. M-OBPE+back and M-OBPE-final are the OBPE with back-translation multilingual models before and after continued training for translation between LRL, respectively. \underline{underline} indicates the model we used for back-translation. \textbf{Bold} represents the best overall model.}
\label{tab:results}
\end{table*}

\begin{table}[t]
\centering
\begin{tabular}{c|S[table-format=3.2]S[table-format=3.2]}
\hline
\textbf{Data} & \textbf{M-OBPE+back} & \textbf{M-OBPE-final}\\
\hline
xho-zul & 1.5 & $\mathbf{11.2}$ \\
zul-sna & 1.9 & $\mathbf{8.8}$ \\
sna-afr & 1.9 & $\mathbf{12.2}$ \\
afr-ssw & 1.3 & $\mathbf{4.9}$ \\
ssw-tsn & 2.0 & $\mathbf{14.5}$ \\
tsn-tso & 2.1 & $\mathbf{13.6}$ \\
tso-nso & 2.4 & $\mathbf{13.2}$ \\
nso-xho & 1.7 & $\mathbf{8.2}$ \\
\hline
avg & 1.8 & $\mathbf{10.8}$ \\
\hline
\end{tabular}
\caption{BLEU scores on Flores dev set for translating between LRLs. M-OBPE+back and M-OBPE-final are the OBPE multilingual models with back-translation before and after continued training for translation between LRL, respectively. M-OBPE-final is the system we submitted for the shared task. \textbf{Bold} represents the best results.}
\label{tab:res-lrl}
\end{table}

\begin{table}
\centering
\resizebox{0.48\textwidth}{!}{\begin{tabular}{c|S[table-format=3.2]S[table-format=3.2]S[table-format=3.2]|S[table-format=3.2]}
\hline
\textbf{Data} & \textbf{BLEU} & \textbf{spBLEU} & \textbf{CHRF2++} & $\mathbf{\Delta} $\textbf{CHRF2++} \\
\hline
sna-eng & 18.7 & 22.1 & 42.9 & 5.5 \\
xho-eng & 24.3 & 26.8 & 47.7 & 5.6 \\
tsn-eng & 19.8 & 22.1 & 42.6 & 7.7 \\
zul-eng & 26.7 & 28.5 & 49.3 & 6.5 \\
nso-eng & 26.5 & 28 & 48.1 & 9.4 \\
afr-eng & 44.7 & 46.4 & 66 & 9 \\
tso-eng & 20.3 & 21.8 & 41.9 & 8.8 \\
ssw-eng & 21.5 & 23.5 & 43.8 & 7.9 \\
\hline
avg & 25.31 & 27.4 & 47.79 & 7.55 \\
\hline
eng-sna & 10.3 & 17.6 & 41.1 & 2.9 \\
eng-xho & 9.4 & 18.6 & 42.5 & 3.4 \\
eng-tsn & 18.8 & 19.7 & 43 & 5 \\
eng-zul & 11.9 & 22.8 & 46.1 & 3.4 \\
eng-nso & 22.7 & 24.1 & 47.8 & 4 \\
eng-afr & 35.9 & 40.5 & 62.2 & 3.6 \\
eng-tso & 15.8 & 17.9 & 41.5 & 4.8 \\
eng-ssw & 7.6 & 15.5 & 38.9 & 4.4 \\
\hline
avg & 16.55 & 22.09 & 45.39 & 3.94 \\
\hline
xho-zul & 8.5 & 18 & 41.4 & 1.9 \\
zul-sna & 8.5 & 15 & 38.7 & 1.7 \\
sna-afr & 12 & 15.1 & 38 & 3.9 \\
afr-ssw & 5.3 & 11.2 & 34.3 & 7.2 \\
ssw-tsn & 14.4 & 15.4 & 38.9 & 2.9 \\
tsn-tso & 13.2 & 15.1 & 38.7 & 2.1 \\
tso-nso & 13.1 & 12 & 36.6 & 5.8 \\
nso-xho & 6.6 & 13.7 & 36.9 & 4 \\
\hline
avg & 10.2 & 14.44 & 37.94 & 3.69 \\
\hline
\hline
overall avg & 17.35 & 21.31 & 43.7 & 5.06 \\
\hline
\end{tabular}}
\caption{The performance of our final system on the shared task test set. $\Delta$ CHRF2++ is the difference between the best submission and our system.}
\label{tab:results2}
\end{table}

\section{Results} \label{sec:res}
We primarily used BLEU score for evaluating all models on the Flores dev set. The final test set evaluation by the shared task organizers additionally used sentence piece BLEU (spBLEU) and chrf2. 

\subsection{Translation Between English and LRLs}
Table \ref{tab:results} shows our results on the translation between English and LRLs. For each translation direction, we selected the best model among the two bilingual models and the 17th epoch checkpoint of the OBPE multilingual to perform back-translation.  Although the multilingual model was trained only for 17 epochs, it outperformed the fully trained bilingual models in some language pairs. Most of these pairs are resource-poor ($eng \leftrightarrow nso, tso, ssw$). The exception of this finding was the translations between English and Afrikaans. These two languages are from the same family, so we hypothesize that the bilingual models did not need help from other resource-rich pairs or additional training examples to translate between the two languages. The training data of resource-richer language pairs ($eng \leftrightarrow xho, zul, tsn$) were sufficient to train good bilingual models.

After we fully trained both OBPE and OBPE+back-translation multilingual models, the OBPE model performed better than the back-translation model in most directions with English as a target language, namely, $sna, tsn, nso, tso, ssw \rightarrow eng$. However, for the three $eng$ generation directions where the back-translation model performed similarly or better than the OBPE model ($xho, zul, afr \rightarrow eng$), the back-translation data was generated from the bilingual models, not the OBPE multilingual model. This synthetic data contains actual English sentences and synthetic LRLs sentences. These translation pairs were relatively resource-rich. In contrast, most of the remaining pairs were resource-poor, and their back-translation data was generated from the partially trained OBPE multilingual model. These results show that although the 17th epoch checkpoint of the OBPE multilingual model was better than bilingual models in resource-poor language pairs, it was not yet good enough for generating text in LRLs. This led to a performance drop for the back-translation model on most of the $eng$ generation directions compared to the OBPE multilingual model.

On the other hand, the back-translation model outperformed the OBPE model in all directions translating into LRLs. These directions require synthetic English sentences and actual LRLs sentences for back-translation. A plausible explanation for this is that learning to translate to English is easier than translating to LRLs for both bilingual and multilingual models.

\subsection{Translation between LRLs}
Table \ref{tab:res-lrl} shows the performance of the OBPE+back-translation model before and after continued training for translation between LRLs. The model’s performance improved on both the initial language pairs (in table \ref{tab:results}) and the new translation directions. Moreover, $sna \rightarrow afr$ and $afr \rightarrow ssw$ were improved using only synthetic data (see section \ref{sec:syn}). We ascribe the success in improving the model’s performance in translating between English and LRLs to the balancing approach (see section \ref{sec:bal}), as we used real training data (not back-translated sentences) in the continued training. 

\subsection{Official Results}
Table \ref{tab:results2} shows the results provided by the shared task organizers for our system as evaluated on a hidden test set. The table also compares the best constrained submission for each translation direction and our system. Our model did not achieve the best performance in any direction. 
However, the teams whose models performed better all trained on all languages included in the shared task (not just Southern African languages).

We hypothesize that this is the main reason for the gap in performance between our system and the better performing ones, as those models could benefit from more training data and increased cross-lingual transfer.
The fact that our model performs relatively worse when translating into English provides some evidence for this: the other systems could benefit learning to translate to English in many more translation directions and with much more data in total.
Given our computational resources, it would have required a total training time of 106 days to cover all language directions in the shared task. Unfortunately this was not feasible in the time provided for the shared task.
The findings paper for the shared task presents more details about other teams’ submissions \citep{adelani-etal-2022-findings}.


\section{Conclusion}

We have presented our multilingual neural MT model for 8 Southern African languages. Until recently, it would not have been possible to train a multilingual model for these languages because of data scarcity. 
During model development we found the benefits of multilingual modelling to be especially great for the lowest-resourced languages. Our results show that overlap BPE, back-translation, and synthetic training data generation are all valuable techniques for low-resource MT.
More generally, we find multilingual modelling to be a fruitful approach to Southern African MT.
For future work we would like to investigate further approaches for training large multilingual models for low-resource languages with a limited compute budget. 

\section*{Acknowledgements}
This work is based on research supported in part by the National Research Foundation of South Africa (Grant Number: 129850).  
Computations were performed using facilities provided by the University of Cape Town’s ICTS High Performance Computing team: \url{hpc.uct.ac.za}.
Francois Meyer is supported by the Hasso Plattner Institute for Digital Engineering, through the HPI Research School at the University of Cape Town.

\bibliography{emnlp2022}
\bibliographystyle{emnlp2022}




\end{document}